\documentclass[10pt,twocolumn,letterpaper]{article}

\usepackage{iccv}
\usepackage{times}
\usepackage{epsfig}
\usepackage{graphicx}
\usepackage{amsmath}
\usepackage{amssymb}

\usepackage{booktabs}
\usepackage[small]{caption}

\usepackage[pagebackref=true,breaklinks=true,colorlinks,bookmarks=false]{hyperref}

\iccvfinalcopy 

\def\iccvPaperID{37} 

\ificcvfinal\fi

\begin{document}

\title{Graph2Pix: A Graph-Based Image to Image Translation Framework}



\author{\stepcounter{footnote}\vspace{1mm}Dilara Gokay $^{1,}$\thanks{Equal contribution. Author ordering determined by a coin flip.}
\hspace{0.75cm}
Enis Simsar $^{1,2,}$\footnotemark[2] 
\hspace{0.75cm}
Efehan Atici $^{2}$
\\
Alper Ahmetoglu $^{2}$
\hspace{0.75cm}
Atif Emre Yuksel $^{2}$
\hspace{0.75cm}
Pinar Yanardag $^{2}$
\vspace{0.1em}
\\
$^1$Technical University of Munich
\hspace{2em} 
$^2$Bogazici University
\\
{\tt\small dilara.goekay@tum.de}
\hspace{0.5em}
{\tt\small enis.simsar@tum.de}
\hspace{0.5em}
{\tt\small efehan.atici@boun.edu.tr}
\\
{\tt\small alper.ahmetoglu@boun.edu.tr}
\hspace{0.5em}
{\tt\small atif.yuksel@boun.edu.tr}
\hspace{0.5em}
{\tt\small yanardag.pinar@gmail.com}
}

\maketitle
\ificcvfinal\thispagestyle{empty}\fi
 
\begin{abstract}
   In this paper, we propose a graph-based image-to-image translation framework for generating images. We use rich data collected from the popular creativity platform Artbreeder\footnote{\url{http://artbreeder.com}}, where users interpolate multiple GAN-generated images to create artworks. This unique approach of creating new images leads to a tree-like structure where one can track historical data about the creation of a particular image. Inspired by this structure, we propose a novel graph-to-image translation model called \textit{Graph2Pix}, which takes a graph and corresponding images as input and generates a single image as output. Our experiments show that Graph2Pix is able to outperform several image-to-image translation frameworks on benchmark metrics, including LPIPS (with a $25\%$ improvement) and human perception studies ($n=60$), where users preferred the images generated by our method $81.5\%$ of the time. Our source code and dataset are publicly available at \url{https://github.com/catlab-team/graph2pix}.
\end{abstract}

\section{Introduction}

Recent breakthroughs in deep learning show that artificial intelligence (AI) can not only be creative, but can also help humans be more creative. Tasks that require human creativity are being taken over by generative approaches as their capabilities increase. Music generation \cite{engel2018gansynth},  face generation \cite{DBLP:journals/corr/abs-1710-10196},   domain translation \cite{CycleGAN}, and image manipulation  \cite{wang2017highresolution} are just a few of the many creative applications of generative models. 

The success of generative methods has also sparked several public tools where researchers and artists  express their creativity through interaction with pre-trained models. For example, the GauGAN platform  is offered as an interactive tool based on an image-to-image translation framework \cite{GauGANPaper} where users design their own lifelike landscape images with properties such as rivers, rocks, grasses, and clouds. Websites like \textit{This Person Does Not Exist}\footnote{\url{https://www.thispersondoesnotexist.com}} use pre-trained GAN models like StyleGAN \cite{StyleGAN} and provide users with uniquely generated faces.

\begin{figure}
    \centering
    \includegraphics[width=0.85\columnwidth]{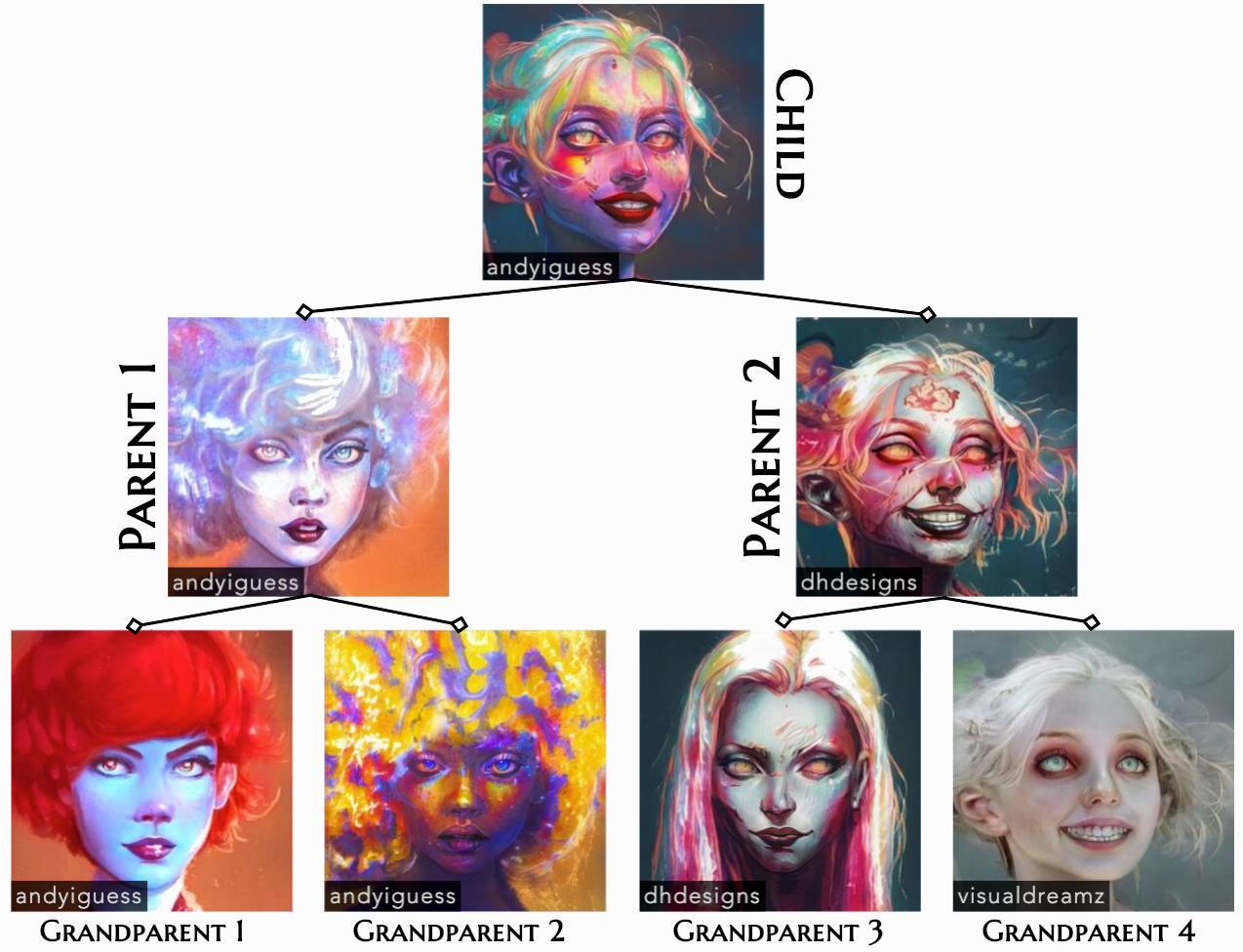}
    \caption[Lineage Footnote]{A sample lineage data up to 2 levels \protect\footnotemark.  The creators of the images are annotated with labels.}
    \label{fig:lineage}
\end{figure}

One of the most popular GAN-based creativity platforms is Artbreeder. The platform's easy-to-use interface has attracted thousands of users, allowing them to create over 70 million GAN-based images. Artbreeder helps users create new images using BigGAN  \cite{BigGAN} or StyleGAN-based  models where users can adjust parameters or mix different images to create new images.  Users can \textit{breed} new images from a single one by editing \textit{genes}  such as \textit{age, gender, ethnicity} or create new ones by \textit{crossbreeding} multiple images together. The unique capability of the crossbreeding function allows the generated images to contain  \textit{ancestry} data where the ancestors of the generated image can be traced in a tree-like structure (see Figure \ref{fig:lineage}).

\footnotetext{Access the full tree via \url{https://www.artbreeder.com/lineage?k=1fcdf872ec11c80e955bb5c1}.}

The lineage structure of images generated with Artbreeder opens up a wide range of possible applications, but also presents unique challenges. The lineage information provides a tree-like structure in which one can trace the parents, grandparents, and further ancestors of a given image. However, it is not entirely clear how such a  structure can be used in GAN-based models. For example, how can we generate a \textit{child} image based on a list of its \textit{ancestors}?  One could try to use image-to-image translation methods such as Pix2Pix \cite{pix2pix} or CycleGAN \cite{CycleGAN} and use any \textit{ancestor} to generate the \textit{child} image (e.g.,  feeding $Parent_{1}$ and generating $Child_{1}$ in Figure \ref{fig:lineage}). However, this approach results in a significant loss of information since only one ancestor can be used (e.g., $Parent_2$ or the grandparents are considered). 

In this paper, we propose a novel image-to-image translation method that takes multiple images and their corresponding \textit{lineage} structure and generates the target image as output. To propose a general solution, our framework takes a graph structure, considering the tree structure of lineage data as a sub-case. To the best of our knowledge, this is the first image-to-image translation method that uses a graph-based structure.

The remainder of the paper is organised as follows.   Section \ref{sec:related_work} discusses related work. Section \ref{sec:methodology} describes  our novel graph-based image-to-image translation framework. Section \ref{sec:experiments} provides details about Artbreeder platform and the collected data, and presents the qualitative and quantitative experiments to demonstrate the effectiveness of our model. Section \ref{sec:limitations} discusses the limitations and failures of our model. Section \ref{sec:conclusion} concludes the paper.

\section{Related Work}
\label{sec:related_work}

\paragraph{Image-to-Image Translation Models}
Generative Adversarial Networks (GANs) are two-part networks that model the real world to the generative space using deep learning methods \cite{NIPS2014_5423}. The generative part of the network tries to create images similar to the dataset while the adversarial part tries to detect if the created image is coming from the training dataset or if it is a generated one. The main goal of GANs is to try to model the image space in a way that makes generated images indistinguishable from the ones in the dataset. Generative networks can generate new realistic images from random noise vectors called latent vectors via internal mapping.

 Image-to-image translation models are usually designed as conditional GAN models, where the goal is to transform one image into the style of another image, retaining some features of the first image and applying others from the second image. They can be used in various applications such as transforming maps into aerial photographs, transforming real images into artist drawings, removing or modifying objects, and image quality enhancement. The first approaches to image-to-image translation problem date back to 2001 \cite{HertzmannImageAnalogies}, where they used more traditional ways like implementing separate, task-specific programs. However, the problem of having domain-specific solutions to a general problem has always been a limiting factor for researchers. The groundbreaking change from task-specific machines to a general-purpose algorithm was made with the discovery of Pix2Pix \cite{pix2pix}. This generalized method for automatically generating lifelike images automatically with setting relevant loss functions for models paved the way for other implementations such as GauGAN \cite{GauGANPaper} and CycleGAN \cite{CycleGAN}. Later, researchers found an impressive solution to image-to-image translation tasks, using a method called U-GAT-IT \cite{kim2019u}. Their approach is to develop a new method for unsupervised learning with attention networks that has normalization layers. They were able to get a neural network to creatively transform images in different styles, such as portraits into anime style, or horses into zebra images. UNIT \cite{liu2017unsupervised} is another unsupervised image-to-image translation network and is built upon the a shared-latent space assumption. This means that UNIT can transform an image from one domain into another and does not require having the corresponding images during training. CC-FPSE \cite{liu2019learning} is tailored for semantic image synthesis. They propose a conditional convolution to efficiently control the generation. TSIT \cite{jiang2020tsit} uses a symmetrical two-stream network to decouple the semantic structure and style information. Recently, pixel2style2pixel \cite{richardson2021encoding} has been shown to perform well in tasks such as multi-modal conditional image synthesis, facial frontalization, and super-resolution.

\paragraph{Graph Convolutional Networks} Graph convolutional networks (GCNs) were introduced by Kipf and Welling \cite{kipf2016semi} and graph structure was used to gain better insight into the hidden representation of data. They showed that GCN usefully incorporates both global (graph level) and local (node level) graph structures for semi-supervised classification. GCN has achieved state-of-the-art results in several application areas, including  social analysis \cite{backstrom2011supervised} and  molecular structural analysis \cite{de2018molgan}. GCNs are also being explored for computer vision \cite{gao2019graph}. Yao \etal \cite{yao2018exploring} proposes a model that combines GCNs with an LSTM architecture to integrate semantic and spatial object relations into an image encoder for image captioning task. Mittal \etal  \cite{mittal2019interactive} proposes a method for generating an image based on a sequence of scene graphs and uses a GCN to handle variable size scene graphs along with generative adversarial image translation networks. Zeng \etal \cite{zeng2019graph} uses GCN to represent proposal-proposal relationships and use it for action classification and localization. DGCNet \cite{zhang2019dual} models two graphs, one representing the relationships between pixels in the image and the other dependency along the channels of the feature map of the network.

 \begin{figure*}[t]
    \centering
    \includegraphics[width=1.85\columnwidth]{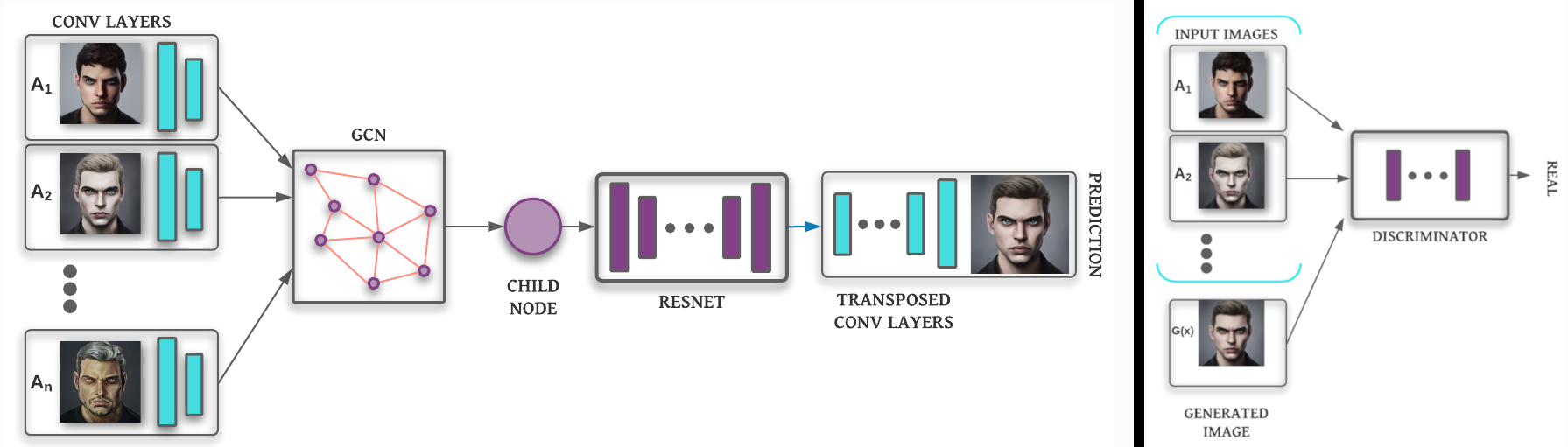}
    \caption{An illustration of Graph2Pix. During the generation process (shown on the left) our 2-layer GCN module takes $A_{1}$ \dots $A_{n}$, the ancestors  of an image as an input, and generates the prediction. The discriminator (shown on the right) takes the concatenation of the input images $A_{1}$ \dots $A_{n}$ and the generated image $G(x)$ and creates a prediction. Note that we placed GCN after the convolutional layer to use it more efficiently.}
    \label{fig:framework}
\end{figure*}

\section{Methodology}
 \label{sec:methodology}
In this section, we first describe a baseline image-to-image translation framework and then explain details of the proposed Graph2Pix model.

\subsection{Pix2Pix and Pix2PixHD}
\label{sec:methodology_part1}
Pix2Pix is an image-to-image translation network based on the conditional GAN structure. Its generator $G$ uses a network structure based on U-Net \cite{U-Net} while its discriminator $D$ uses a  convolutional classifier called Patch-GAN \cite{Patch-GAN}. It takes a pair of images $\{(y_i, x_i)\}$ as input (e.g., a label map and the corresponding image) and computes the conditional distribution of the images via $\min_{G}\max_{D} \mathcal{L}_{cGAN} (G, D)$.
 
Pix2PixHD \cite{Pix2PixHD} is built upon the Pix2Pix structure to generate higher resolution images. Compared to Pix2Pix, it divides the generator network $G$ into two separate parts: the global generator network $G_1$, and  local enhancer network $G_2$. Although they are similar in concept, $G_1$ is trained on the down-sampled versions of the  training images while $G_2$ acts as a wrapper around $G_1$ to increase the image resolution of its low-resolution output. This network design supports scalability. To generate higher resolution images, additional generator networks $G_3 \dots G_n$ can be added. Pix2PixHD also improves the discriminator component with a multi-scale approach where each image scale level has a separate discriminator, $D_1 \dots D_3$, respectively. As the images are progressively down-sampled, the discriminator is be able to capture the global consistency between real and fake images.

Our proposed model adopts the architecture of Pix2PixHD as a backbone and extends it with a multi-generator that fuses images using a graph convolutional network, which we introduce below.

\subsection{Graph2Pix}
 
Pix2PixHD is not suitable for our task because it is limited to a single source and cannot accommodate multiple sources. Therefore, it does not take into account the lineage structure of the input since it cannot use the ancestral information such as \textit{parents} or \textit{grandparents} of the \textit{child} image (see Figure \ref{fig:lineage}). In contrast, a GCN-based approach offers the potential to capture the relationship among the members of a family, which can be represented as a graph.

Graph-based neural networks use a similarity graph that allows the information propagation between similar inputs.  In this work, we use a GCN  \cite{kipf2016semi} that allows CNNs to work directly with graphs, where each layer in GCNs updates a vector representation of each node in the graph by considering the representations of its neighbors. The forward computation of a graph convolutional layer is as follows:
\begin{equation}
    H^{(l+1)} = \sigma(\Tilde{D}^{-\frac{1}{2}}\Tilde{A}\Tilde{D}^{-\frac{1}{2}} W^{(l)} H^{(l)})
\label{eq:gcn}
\end{equation}
where $\Tilde{A}$ is the adjacency graph including self-loops, $\Tilde{D}$ is a diagonal matrix that contains the degree of node $i$ in $\Tilde{D}_{ii}$, and $\sigma$ is a non-linear activation function. Here, $W^{(l)}$ is a trainable weight matrix in layer $l$, and $H^{(l)}$ represents the input activations in layer $l$. $\hat{A}=\Tilde{D}^{-\frac{1}{2}}\Tilde{A}\Tilde{D}^{-\frac{1}{2}}$ can be viewed as a normalized adjacency graph. A graph convolutional layer differs from a conventional fully connected layer in the sense that it combines information from different nodes in each layer. This operation turns out to be a first-order approximation of spectral convolutions, which is why the method is called graph convolutional networks \cite{kipf2016semi}.

Our proposed global generator consists of a convolutional front-end, a set of residual blocks, and a transposed convolutional back-end (see Figure \ref{fig:framework}). Graph convolutional layers are inserted between the convolutional block and the residual block. The convolutional layers create low-resolution representations of the images in the graph. These low-resolution representations are inputs to the graph convolutional layers. Since we are working in the image domain, we replace the affine transformations in GCN with $3 \times 3$ convolutions. Then, the graph convolutional layer computes:
\begin{equation}
    H^{(l+1)} = \sigma(\Tilde{D}^{-\frac{1}{2}}\Tilde{A}\Tilde{D}^{-\frac{1}{2}}\textnormal{Conv}(H^{(l)}))
\end{equation}

Since we are trying to generate the child image based on its ancestors, we cannot provide the child image as input to the pipeline. However, we want to retain the adjacency information, i.e., the graph, to generate the output with respect to the ancestors. To achieve this, we replace the child image with noise, use the noise as input instead, and predict its original version. This procedure is analogous to denoising autoencoders \cite{vincent2008extracting}, where we randomly noise the input features. In the end, the network should faithfully reconstruct the child image based on its ancestors. We also noise random nodes in the graph, instead of the child node, to force the model to learn relationships in the graph. For example, if we noise a parent node in the graph, the model should predict that node by combining information from the grandparent node and the child node. Since the global generator exploits the graph structure of the lineage, it is important to have a sufficiently large lineage. Therefore, we only took the children that have at least two parents and at least one grandparent from those two different parents. Thus, the lineages we used had at least four ancestors.

Similar to the global generator, Pix2PixHD's multi-scale discriminator is not tailored to process multiple source images as it only considers a single input image. Instead, we used the concatenation of all input images as input to the multi-scale discriminator (see Figure \ref{fig:framework}). In the next section, we show that our graph-based approach performs better than the state-of-the-art image-to-image translation methods both quantitatively and qualitatively.

\section{Experiments}
\label{sec:experiments}
In this section, we first describe our dataset and present our experimental setup. We then present the results of our quantitative and qualitative experiments.
  
\begin{figure} 
    \centering
    \hspace{-0.4cm}
      \centering
      \includegraphics[width=0.8\columnwidth]{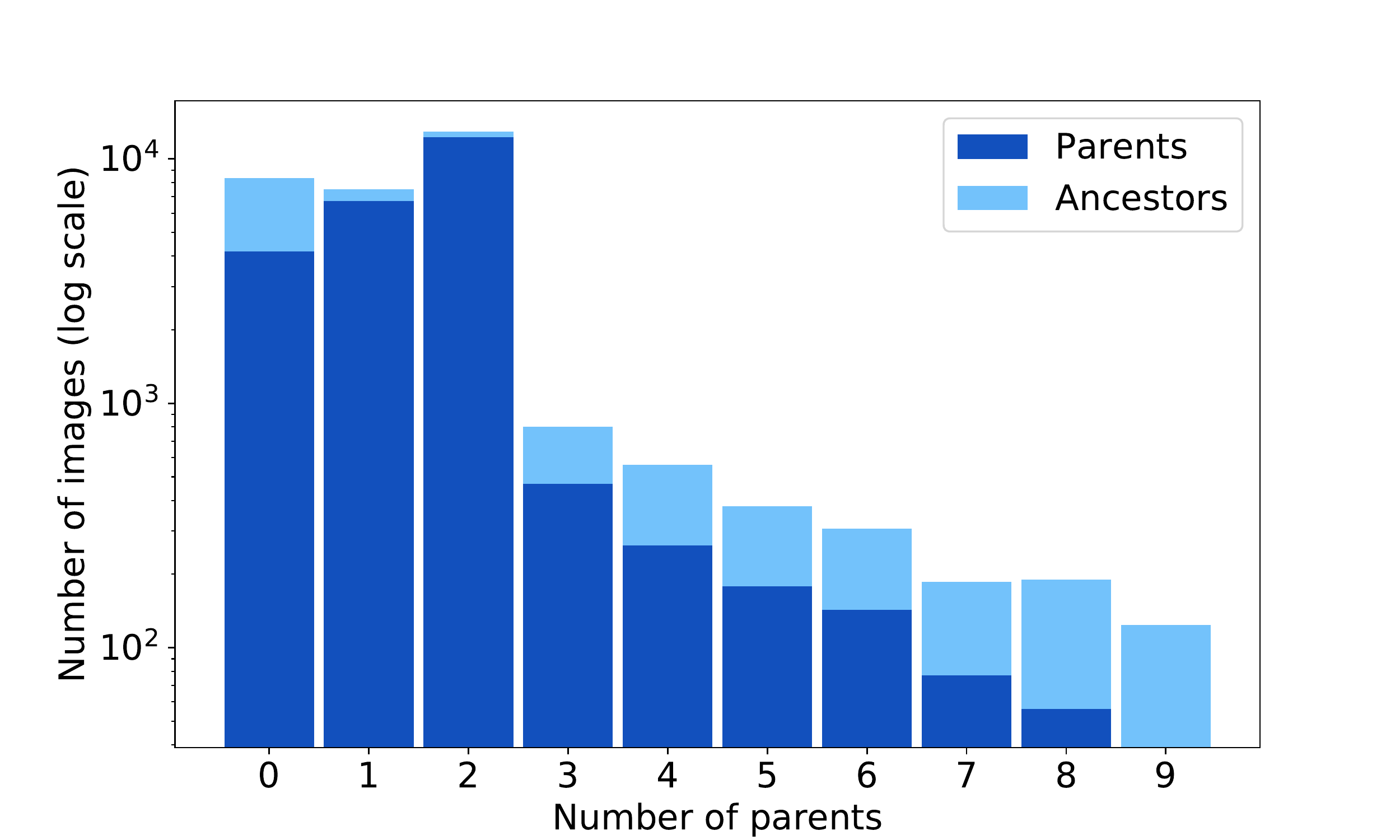}
    \caption{The distribution of parents and ancestors up to 10 is shown. As can be seen from the figures, the majority of the images have  parents and ancestors between 1-4.}
    \label{fig:parents_ancestors}
\end{figure}
 
\begin{figure*}[t]
    \centering
    \includegraphics[width=1.85\columnwidth]{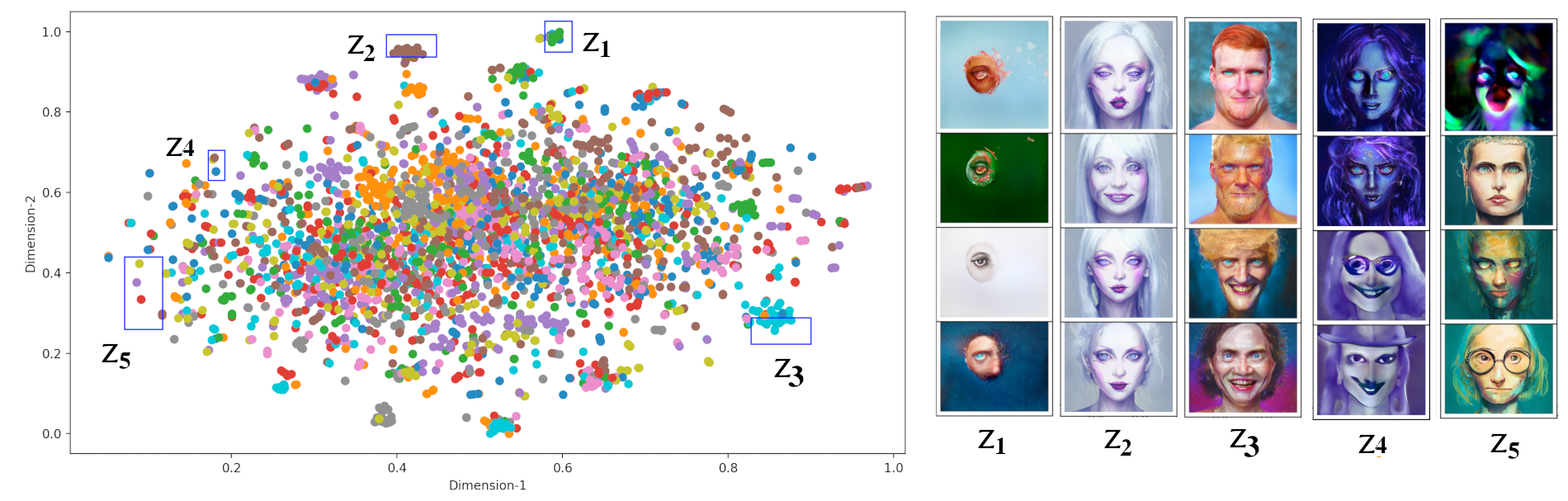}
    \caption{Exploration of the generated images based on user groups. The images are mapped into the 2-dimensional space while maintaining their proximity through t-SNE \cite{TSNE}. The color of the points denotes the user who created the image. The clusters labeled with $Z_{1}$ \dots $Z_{5}$ on the left show selected image groups whose corresponding images are shown on the right.} 
    \label{fig:tsne}
\end{figure*}

\subsection{Dataset}
Artbreeder platform offers two ways to generate images: \textit{mutation} and \textit{crossbreed}. The \textit{mutation} option allows users to create a new image by editing various traits, called \textit{genes}, and allows users to directly modify the original images. Users can change high-level features such as \textit{age} and \textit{gender}, as well as low-level features such as the \textit{red, green,} and \textit{blue} channels of the image. An alternative method is called crossbreeding,  which allows users to mix multiple images.  In this paper, we are only interested in images created by \textit{crossbreeding}. We collected the images from Artbreeder \footnote{Note that the images on Artbreeder are public domain (CC0) and we have obtained permission from the owner to crawl the images.} periodically by crawling the \textit{Trending} section of the `Portraits` category daily over a four-week period. The resolution of the images offered by Artbreeder varies from $512 \times 512$ to $1024 \times 1024$. To ensure that we have a consistent dataset, we have resized all images to $512 \times 512$.  For each image, we collect the `lineage' information that allows us to track the chronological creation of the image. The lineage information includes the parent image, the creation date, and the creator's name. We collected a total of 101,612 images. Figure \ref{fig:parents_ancestors} illustrates the distribution of ancestors and parents of a child node. Although the number of parents in the figures reaches up to $10$ for clarity, the majority of child images have an average of $20.7$ ancestors and $1.6$  first-degree parents.

\begin{figure}
    \centering
    \hspace{-0.4cm}
      \centering
      \includegraphics[width=0.8\columnwidth]{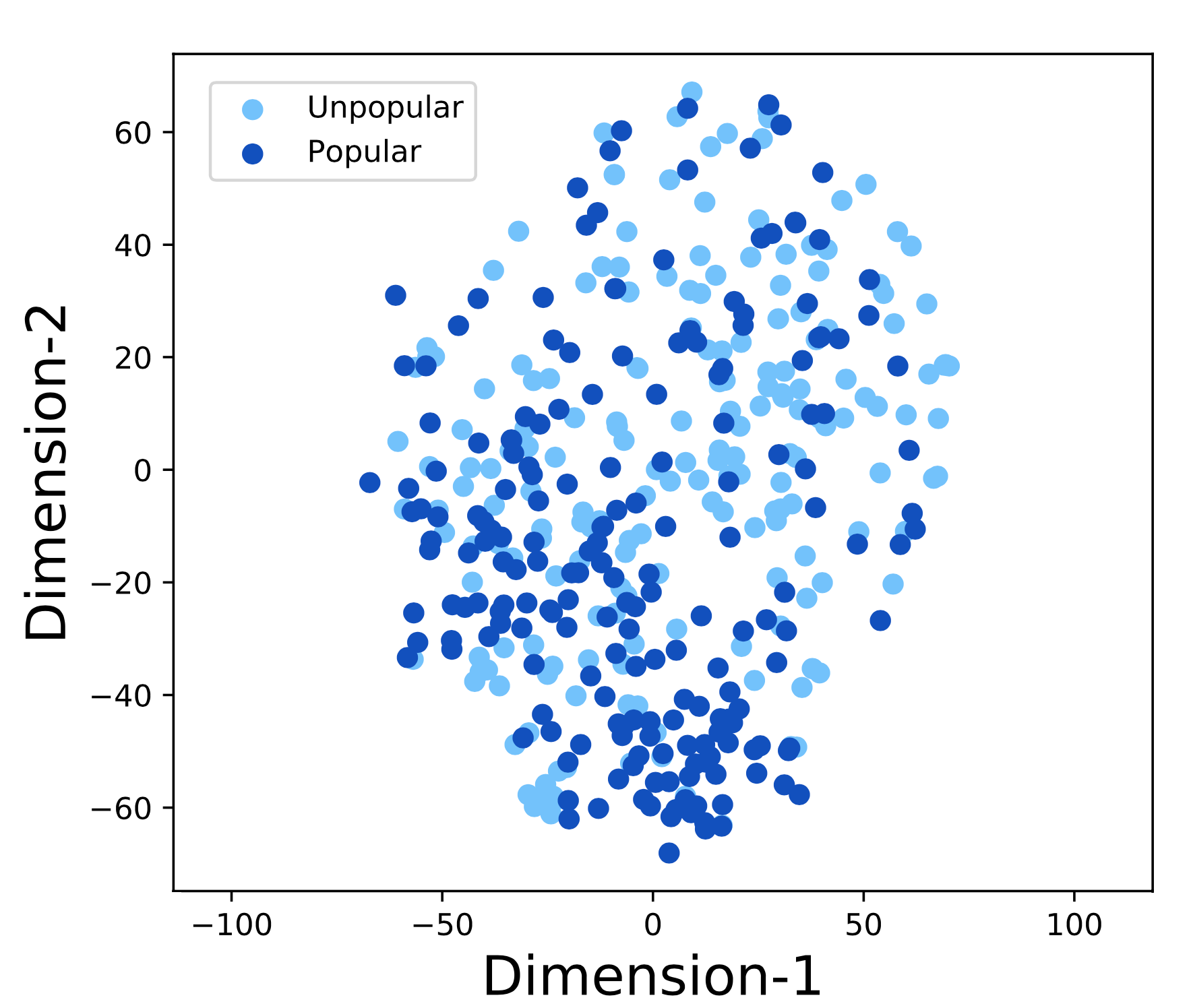}
    \caption{A comparison of popular and unpopular images in 2-dimensions using the t-SNE approach.}
    \label{fig:pop_unpop}
\end{figure}

\paragraph{Data exploration.}  To gain insight into the creative crowd behind Artbreeder, we visually explore the latent space of images and users. We obtain the latent vectors provided by Artbreeder and project each vector as a point in 2-dimensional space. To achieve this, we first reduced the dimensionality of the latent vectors from 9126 dimensions to 50 by performing principal component analysis (PCA) \cite{PCAAnalysis}. After reducing the dimensions of the latent vectors, we apply t-Distributed Stochastic Neighbor Embedding \cite{TSNE} for two-dimensional representation, preserving the proximity in the high-dimensional space. 

t-SNE is an algorithm based on Stochastic Neighbor Embedding \cite{SNE} that aims to visualize high-dimensional data by embedding it in a low-dimensional space. It is tailored to model each data point through a two- or three-dimensional representation such that similar entries are closer together in the final visualization, while different entries are more likely to be farther apart. 

To explore the relationship between users in latent space, we used t-SNE to visualize images. We first extracted embeddings learned from a deep neural network \cite{NeuralNetworkVisualization, VisualizingRepresentationsDeepLearningandHumanBeingscolahsblog-2020-06-02} and used Euclidean distance to compute the distance between the PCA-reduced embedding vectors using the Barnes-Hut algorithm \footnote{Learning rate at 200, number of iterations at 1000}. Figure \ref{fig:tsne} shows a set of selected image groups with their corresponding visual content, where the colors represent different users. For clarity, only a subset of users who created between 50 and 100 images are shown. While the center of the figure shows that similar images can be created by different users, it can also be observed that certain users have a certain style that distinguishes them from others (such as $Z_{1}$, $Z_{2}$ and $Z_{3}$). On the other hand, different users with similar styles can also be observed in different clusters, such as $Z_{4}$ and $Z_{5}$. 

\begin{figure*}
    \centering
    \includegraphics[width=1.85\columnwidth]{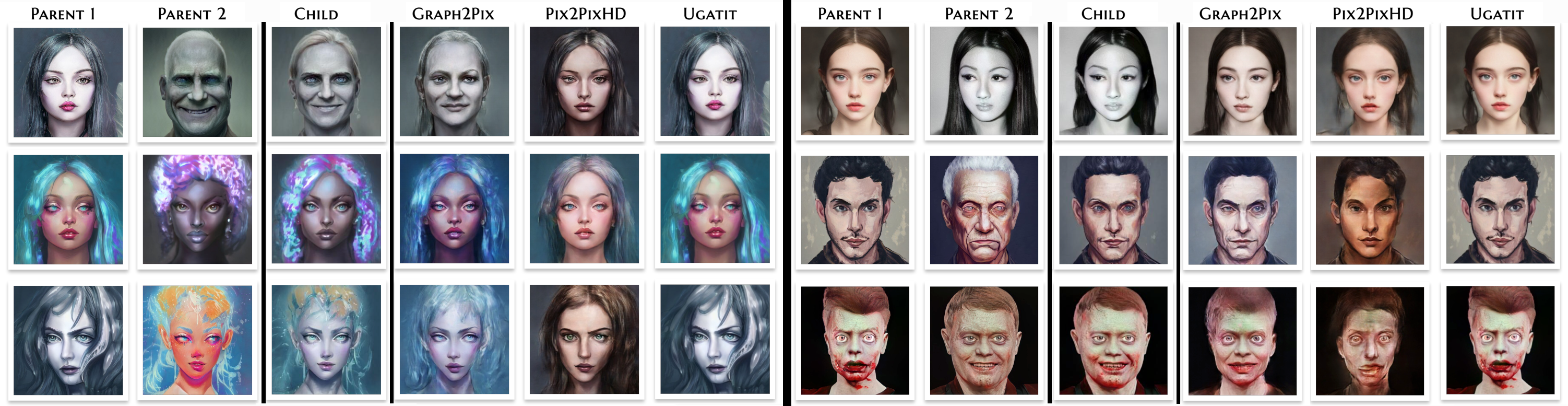}
    \caption{A qualitative comparison of our method with the top performing competitors, Pix2PixHD and U-GAT-IT. First-level ancestors ($Parent_1$ and $Parent_2$) are shown on the left where the ground truth image is denoted with $Child$. As can be seen from the results, our method is able to incorporate the ancestor information when generating the images whereas Pix2PixHD and U-GAT-IT are limited to a single ancestor image ($Parent_1$). Due to lack of space, only first-degree parents are  shown, however a sample lineage for the bottom-left image can be seen on Artbreeder. \protect\footnotemark} 
    \label{fig:results}
\end{figure*}

The Artbreeder platform has a \textit{Trending} page that showcases the images currently most used by users. To understand the relationship between popular and unpopular images, we selected the top 200 users  whose images are used the most, and 200 unpopular users whose images are used the least. Similar to our previous experiment, we used the t-SNE approach and projected the users onto a two-dimensional space after reducing their embedding vectors using PCA. Figure \ref{fig:pop_unpop} illustrates popular and unpopular images. As can be seen from the figure, unpopular images are scattered over a large area, while popular images seem to be concentrated in a specific region (lower left and bottom), suggesting that they share similar features that appeal to a large number of users.

\paragraph{Experimental setup.} We train our model with 86,373 training images and evaluate it with a test set of 15,239 images. We  used 8 as the batch size to generate images with a resolution of  $256 \times 256$. While our method requires the ancestor images up to two levels and the corresponding graph structure as input, the competing methods are limited to a single input (e.g., one of the first level parent images). However, since the results can be sensitive to which parent image is used as input, we run all the competing methods twice, once for each parent, and report the average results. For our experiments, we used the PyTorch platform, and each model was trained using its official implementation \footnotetext{{\url{https://www.artbreeder.com/lineage?k=f8c31131db9a73ac0425}}} \footnote{\url{https://github.com/junyanz/pytorch-CycleGAN-and-pix2pix}, \url{https://github.com/NVIDIA/pix2pixHD}} with the default parameters.  All methods are trained for 50 epochs on a Titan RTX GPU, and U-GAT-IT method is run for 100K iterations since it has no epoch option. 

\paragraph{Baselines.} We compared our performance against several state-of-the-art image-to-image translation frameworks, including Pix2PixHD \cite{Pix2PixHD}, Pix2Pix \cite{pix2pix}, CycleGAN \cite{CycleGAN}, MS-GAN \cite{mao2019mode}, and U-GAT-IT \cite{ugatit}.  Pix2PixHD and Pix2Pix are supervised models that require paired data, as explained in Section \ref{sec:methodology_part1}. CycleGAN is one of the most popular image-to-image translation methods and uses a cycle consistency loss to enable training with unpaired data, where mapping an image from the source domain to the target domain and repeating the process in reverse leads the same starting image. However, as we will show later in our experiments, while CycleGAN generates high quality images, the cycle consistency constraint enforces a strong link between domains and does not learn to transform the input parent image into the output child image, but learns to output the parent image itself. A similar trend can be observed for another unsupervised model, U-GAT-IT, which includes an attention module to focus on more important regions that distinguish between source and target domains. We also compare with  MS-GAN, which uses a mode seeking regularization term that explicitly maximizes the ratio of the distance between the generated images and the corresponding latent codes. We used the Pix2Pix-Mode Seeking option of MS-GAN. 

Note that some of the popular image-to-image translation models are not applicable to our experiments because they specifically require a label map that does not exist in our dataset, such as SPADE \cite{GauGANPaper}, or they require the dataset to contain multiple categories during training, such as FUNIT \cite{liu2019few}. 

It should also be noted that the child images on Artbreeder are not simply an average of two parent images: These images are interpolated according to a specific ratio chosen by the user. Therefore, one might consider estimating the synthesis parameters of Artbreeder directly, rather than developing a new method, as a more relevant baseline. Although this would have been an interesting comparison, we would like to emphasise that Artbreeder platform does not keep the synthesis parameters on its servers. Therefore, this information is not available for our experiments. Thus, using a GCN model becomes useful as it implicitly learn how the ancestors are mixed without explicit ratio data.

\begin{figure*}
    \centering
    \includegraphics[width=1.85\columnwidth]{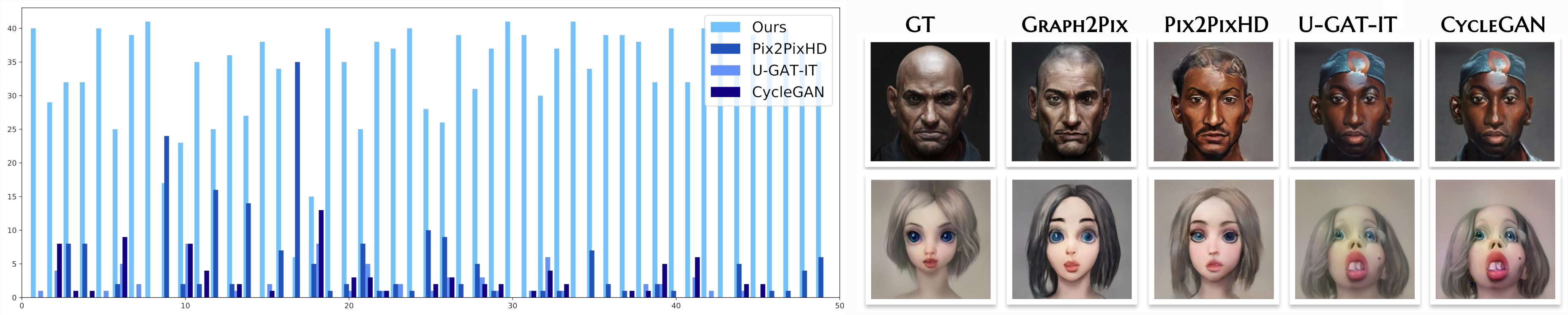}
    \caption{The results for each question in our human evaluation survey are shown on the left  (the x-axis represents the image IDs, the y-axis represents the number of votes received for each method). The top-performing image, where participants unanimously found our method was most successful, is shown in the upper right. Our method is the least successful on the image shown on the bottom right, where participants preferred the Pix2PixHD result. } 
    \label{fig:histogram}
\end{figure*}
\subsection{Quantitative Evaluation}
\label{sec:quantitative}
 
 For our quantitative experiments, we used Learned Perceptual Image Patch Similarity (LPIPS) \cite{zhang2018unreasonable}, Fréchet Inception Distance (FID) \cite{FIDHeusel}, and Kernel Inception Distance (KID) \cite{KIDmetric}. LPIPS uses features from an ImageNet pre-trained AlexNet and measures the diversity of the generated images using the L1 distance. FID is another method for computing image quality and compares the generated samples with the real images in the dataset using Inception-V3 model \cite{InceptionV3}. KID is a more recent metric that is similar to FID but is calculated by finding the Maximum Mean Discrepancy \cite{MMD} between Inception activation layers. One of the advantages of KID compared to FID is that KID compares the skewness of the distributions as well as the mean and variance, and has a faster and unbiased estimator. 
 
Among these metrics, we argue that LPIPS is better suited to evaluate the performance of our system because it computes the score for corresponding image pairs. In other words, it considers the relationship between the ground truth and the generated image, while FID and KID scores are computed using mean and variance statistics on the entire set of images instead of pairwise comparison. The comparison is shown in Table \ref{table:results}. Our method significantly improves the LPIPS metric (lower score is better) compared to its competitors. Our method achieves an LPIPS score of $0.25$, while the second best score is achieved by Pix2PixHD with $0.32$. This result shows that our method is able to improve the LPIPS metric by almost $25\%$. The second metric we use to compare different methods is the FID score, where lower is better. The lowest FID score is obtained by our model, albeit with a slight improvement. This result indicates that both our method and its closest competitors, CycleGAN and U-GAT-IT, are able to generate realistic results. However, as we will show later in our qualitative experiments ( Section \ref{sec:qualitative}), both methods fail to generate images that resemble ground truth. Instead, they only learn to copy the input image (in this case, one of the parent images). On the other hand, we note that our method significantly improves the FID score of the baseline Pix2PixHD method by $19.25\%$. Our method also outperforms its competitors in terms of KID score, which compares the skewness of the distributions and the mean and variance statistics.

\paragraph{Ablation Study.}  In addition to comparing our method to existing methods, we investigated whether using a GCN helps the learning process as follows. We created a random adjacency matrix and trained our model without using any lineage information. When we compare the generated images of the random adjacency based model with our model, we observe that FID and LPIPS scores degraded by 53.17\% and 37.84\%, respectively. This ablation study verifies that our network indeed learns the underlying topology information.

\begin{table}
\centering
  \begin{tabular}{lllllllll}
    \toprule
Method &  LPIPS$\downarrow$  & FID$\downarrow$  &   KID$\downarrow$   & Human$\uparrow$  \\
    \midrule
    
    Pix2PixHD & 0.32   &  23.89   &   0.71   &0.097      \\
    CycleGAN & 0.37     & 19.35   &    1.36   & 0.043   \\ 
    U-GAT-IT & 0.33  &  19.35  & -1.00  &  0.044     \\
        MS-GAN &  0.46    & 85.62  &    5.18 &   N/A    \\
        Pix2Pix & 0.52    &  112.01  &  9.73   &N/A       \\\hline
  \textbf{Graph2Pix} & \textbf{ 0.25}   & \textbf{19.29} &      \textbf{ -1.12 }  & \textbf{ 0.815 }   \\

  \bottomrule
\end{tabular}
\caption{Comparison of our method with baselines in quantitative and qualitative experiments. For LPIPS, FID and KID lower is better. For human evaluations (denoted by \textbf{Human}), higher is better.}
\label{table:results}
\end{table} 

\subsection{Qualitative Evaluation}
\label{sec:qualitative}

In this section, we first present randomly selected images and compare them with the top two competitors, Pix2PixHD and U-GAT-IT. For clarity, we omit the CycleGAN images from the comparison, as it generates images that are very similar to those of U-GAT-IT (see Figure \ref{fig:histogram} for a comparison between the results of CycleGAN and U-GAT-IT). Figure \ref{fig:results} lists the two parent images, the child image (i.e., the \textit{ground truth}), and the images generated by Graph2Pix, Pix2PixHD, and U-GAT-IT, respectively. To allow consistent comparison, we only feed the first parent ($Parent_1$) to the competitors to assess how well they were able to use this information. The first observation is that U-GAT-IT generates almost identical images to the parent and does not learn how to transform from the parent image domain to the child image domain. We think this is due to the unsupervised nature of the model. On the other hand, we can see that Pix2PixHD is able to generate images that are different from the parent, but it fails to generate images that are close to the ground truth. Since our method is able to consider both the parent and its ancestor, it achieves a better result that is closer to the ground truth.

We further conducted a qualitative analysis with 60 human participants and asked them to compare our method with the baseline approaches. Our experimental setup follows the standard approach employed by popular image-to-image translation methods \cite{choi2020stargan,Pix2PixHD}, where participants are given a pair of source and reference images and instructed to choose between candidate images. For brevity, we selected the top 3 competitors according to our quantitative analysis in Section \ref{sec:quantitative}, namely Pix2PixHD, U-GAT-IT, and CycleGAN. We randomly select 50 images for our experiment. For each image, we show the ground truth as the reference image, followed by four images (generated by our method and the competitors) whose order is randomly shuffled, and we ask which image is most similar to the reference image. Participants took an average of 9 minutes to complete the survey. As shown in Table \ref{table:results}, our method obtains the majority of votes ($81.5\%$), followed by Pix2PixHD ($9.7\%$).

We also examined the votes on each image in Figure \ref{fig:histogram}. As can be seen from the figure, our method consistently receives the majority of votes on all images with a few exceptions. It is noticeable that our method is unanimously chosen as the best approach for some images, such as the top right image in Figure \ref{fig:histogram}, where we can see that the skin color and facial structure are much more similar to the ground truth than the competitors. This example shows that the ancestor information is very important for some images and including this structure adds value to the image generation process. In contrast, our method performed poorly on some images, such as the image on the bottom right of Figure \ref{fig:histogram}. 

\begin{figure}
    \centering
    \includegraphics[width=0.85\columnwidth]{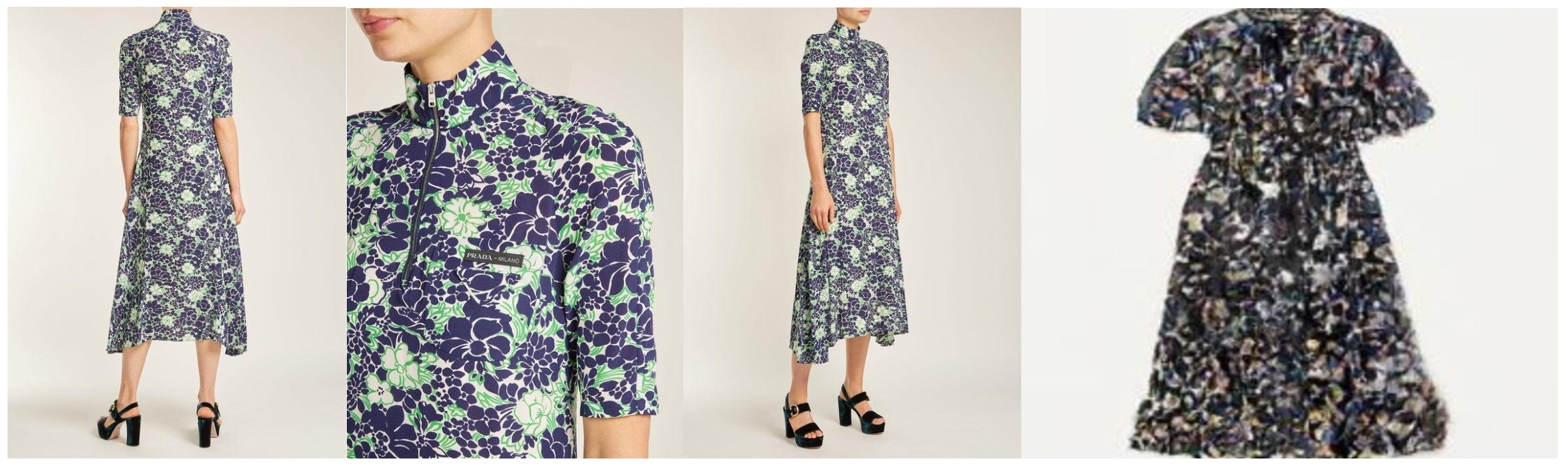}
    \caption{The images in the first three columns are inputs, while the image in the last column is generated by our framework. All input images are treated as a first-degree parent, and the expected cloth image in this case is the child in the lineage.} 
    \label{fig:product}
\end{figure}

\subsection{Further Use Cases}
Our method is not limited to generating images from Artbreeder. Due to the flexibility of the graph structure, our method is also applicable to any scenario where multiple images need to be translated into one, such as multi-source human image generation \cite{lathuiliere2020attention}. To show the validity of our method in different use cases, we applied our method to this case as well. We used the images of a cloth on a model from three different views (front, back and close) as input and generated the image of the cloth itself \footnote{Source: \href{https://modesens.com/product/prada-asymmetric-floralprint-stretchsilk-midi-dress-navy-9969054/}{https://modesens.com/product/prada-asymmetric-floralprint-stretchsilk-midi-dress-navy-9969054/}} (Figure \ref{fig:product}). To see if having multiple parents are indeed helpful, we applied our framework to an image from a single view (front) and compared it to the multiple view scenario based on FID score. We obtained $45.26$ in the single view case, while obtaining $43.43$ in the multiple view case. Moreover, the comparison between these scores and FID score in Table \ref{table:results} shows that the network learned Artbreeder dataset better, implying that a deeper lineage structure improves the performance of our graph-based image-to-image translation framework.

\section{Limitations and Failure Cases}
\label{sec:limitations}
\begin{figure}
    \centering
    \includegraphics[width=0.9\columnwidth]{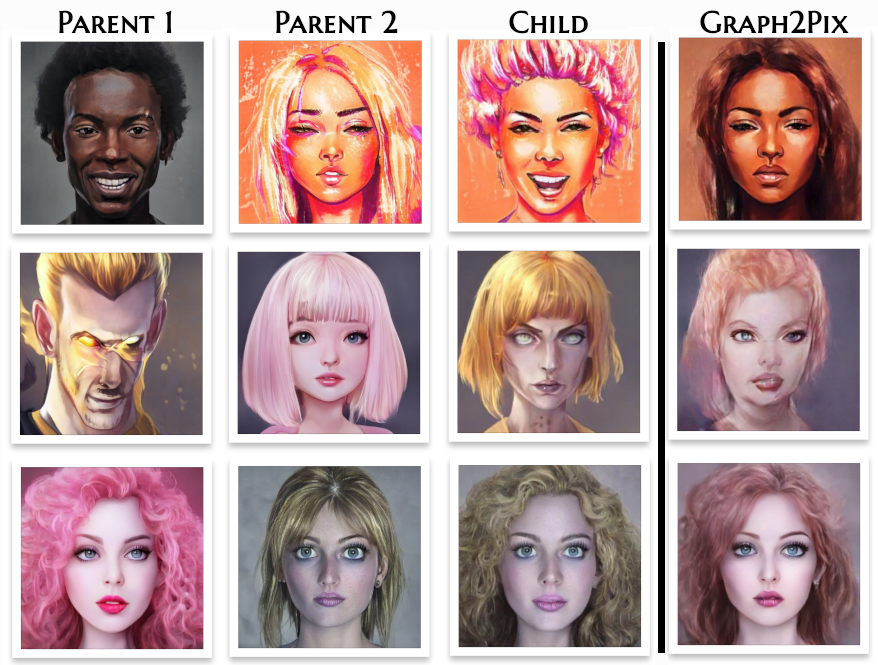}
    \caption{Some cases where our method fails in learning coloring when the parent images have extreme differences.} 
    \label{fig:failure}
\end{figure}

In Figure \ref{fig:failure}, we show some of the failure cases of our method. It is noticeable that although our method is able to take into account the ancestral information and generate images that resemble the ground truth, it fails when the colors of the parent images vary significantly. Another reason for this effect could be that some users \textit{crossbreed} images with different ratios. Therefore, the child image may carry more features from one of the parents (like the top image in Figure \ref{fig:failure}) and our model is not able to capture such dramatic ratios.

An important limitation of our method is that the input images should have a graph- or tree-based structure. However, we argue that due to the flexibility of the graph structure, our method is also applicable to any scenario where multiple images need to be translated into one, such as human image generation  \cite{lathuiliere2020attention} from multiple sources. In addition, we believe that our work can be explored in 3D reconstruction settings or in multi-agent communication settings where multiple drones share their image captures from different locations and construct a single coherent image (see
\cite{mitchell2020gaussian} for a similar setting).
\section{Conclusion \& Future Work}
\label{sec:conclusion}
 
 In conclusion, we propose Graph2Pix, an image-to-image translation method that considers multiple images and their corresponding graph structure as input and generates a single image as output. To the best of our knowledge, our method is the first to propose a GCN-based image-to-image translation method. Our quantitative experiments show that our method outperforms the competitors and improves LPIPS, FID, and KID scores by $25\%$, $0.3\%$, and $12\%$, respectively. Furthermore, our qualitative experiments show that human participants prefer the images generated by our method $81.5\%$ of the time compared to its competitors. For future work, we consider using metadata such as popularity as a way to indicate the importance of the ancestors and we plan to extend our GCN module to capture hierarchical graphs. In addition, we plan to extend our method to other multi-source problems, such as generating human images from multiple sources.
  \paragraph{Acknowledgments.} This publication has been produced benefiting from the 2232 International Fellowship for Outstanding Researchers Program of TUBITAK (Project No: 118c321). We also acknowledge the support of NVIDIA Corporation through the donation of the TITAN X GPU and GCP research credits from Google. We also would like to thank Joel Simon for their support in collecting the dataset. 
  
\newpage 

{\small
\bibliographystyle{ieee_fullname}
\bibliography{egbib}
}

\end{document}